\pdfoutput=1

\documentclass[11pt,a4paper]{article}
\usepackage[hyperref]{ranlp2023}
\usepackage{times}
\usepackage{latexsym}

\aclfinalcopy

\usepackage{microtype}

\author{Nilay Patel \\
  University of California, Santa Cruz \\
  \texttt{nilay@ucsc.edu} \\\And
  Jeffrey Flanigan \\
  University of California, Santa Cruz \\
  \texttt{jmflanig@ucsc.edu}}
\date{\today}
\title{Forming Trees with Treeformers\jmf{maybe change to something that includes compositional generalization}}

\usepackage{graphicx}
\usepackage{amsmath, amsthm, amssymb}
\usepackage{booktabs}
\usepackage{tcolorbox}
\usepackage{tikz}
\usepackage{colortbl}
\usepackage{algorithm}
\usepackage[noend]{algpseudocode}
\usetikzlibrary{trees,positioning}

\newcommand{\OO}{\mathcal{O}}
\renewcommand{\bold}{\textbf}
\definecolor{mblue}{rgb}{0.05,0.49,0.73}  % light
\definecolor{mred}{rgb}{0.79,0.18,0.22}   % light
\definecolor{mpurple}{rgb}{0.5, 0, 0.5}
\definecolor{verylightgray}{rgb}{0.85, 0.85, 0.85}
\definecolor{highlight}{rgb}{0.66, 0.85, 0.02}

\newcommand{\jmf}[1]{} %uncomment to remove comments
\newcommand{\nilay}[1]{}
\newcommand{\gr}[1]{}
\newcommand{\bdk}[1]{}

\begin{document}

\maketitle

\begin{abstract} 
Human language is known to exhibit a nested, hierarchical structure, allowing us to form complex sentences out of smaller pieces. However, many state-of-the-art neural networks models such as Transformers have no explicit hierarchical structure in its architecture---that is, they don't have an inductive bias toward hierarchical structure. Additionally, Transformers are known to perform poorly on compositional generalization tasks which require such structures. In this paper, we introduce Treeformer, a general-purpose encoder module inspired by the CKY algorithm which learns a composition operator and pooling function to construct hierarchical encodings for phrases and sentences. Our extensive experiments demonstrate the benefits of incorporating hierarchical structure into the Transformer and show significant improvements in compositional generalization as well as in downstream tasks such as machine translation, abstractive summarization, and various natural language understanding tasks.

%and are used in many of the latest pre-trained models. Transformers use tokens as the unit of information. That is, each token is encoded into a vector representation, and those vectors are used directly in a computation. However, humans frequently consider spans of tokens (i.e., phrases) instead of their constituent tokens. In this paper, we introduce Treeformer, an encoder module inspired by the CKY algorithm and Transformer which learns a composition operator and pooling function in order to construct hierarchical encodings for phrases and sentences. Our extensive experiments demonstrate the benefits of incorporating hierarchical structure into the Transformer, and show significant improvements compared to the Transformer model in machine translation, abstractive summarization, and various natural language understanding tasks.
\end{abstract}

\section{Introduction}%
\label{sec:introduction}
%\jmf{make clear that Treeformer layer is not a model by itself}

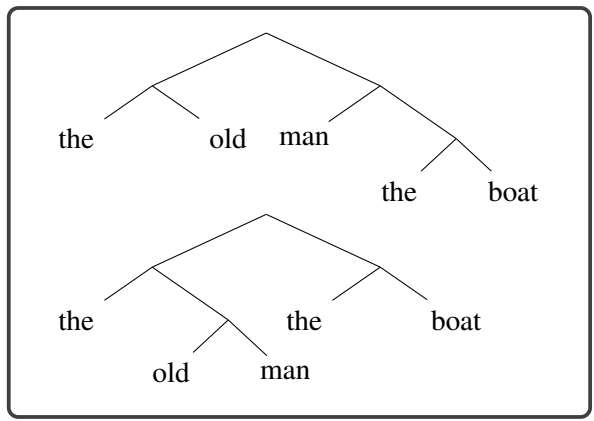
\begin{figure}[t]
  \centering
  \begin{tcolorbox}[minipage,arc=0pt, standard jigsaw, opacityback=0, outer arc=0pt, size=normal]
    \begin{tikzpicture}
      \tikzstyle{level 1}=[sibling distance=30mm, level distance=7mm]
      \tikzstyle{level 2}=[sibling distance=20mm, level distance=7mm]
      \tikzstyle{level 3}=[sibling distance=15mm, level distance=7mm]
      \coordinate (a)
      child { coordinate (b)
          child {node (c) {the}}
          child {node (d) {old}}
        }
      child { coordinate (e)
          child {node (f) {man}}
          child { coordinate (g)
              child {node (h) {the}}
              child {node (i) {boat}}
            }
        };
    \end{tikzpicture}
    \begin{tikzpicture}
      \tikzstyle{level 1}=[sibling distance=30mm, level distance=7mm]
      \tikzstyle{level 2}=[sibling distance=20mm, level distance=7mm]
      \tikzstyle{level 3}=[sibling distance=15mm, level distance=7mm]
      \coordinate (a)
      child { coordinate (b)
          child {node (c) {the}}
          child {coordinate (d)
              child {node (e) {old}}
              child {node (f) {man}}
            }
        }
      child { coordinate (g)
          child {node (i) {the}}
          child {node (j) {boat}}
        };
    \end{tikzpicture}
  \end{tcolorbox}
  \caption{Two different parses of the text ``the old man the boat'' with significantly distinct
    meanings. While the top parse is a complete sentence (with ``man'' as a verb), the second is
    nonsense. Therefore, the encodings for the subphrase ``the old man'' (for example) in these parses
    should be significantly different.}%
  \label{fig:different-parses}
\end{figure}

Human language is known to exhibit a nested, or hierarchical structure \citep{chomsky,montague}.  This structure allows humans to construct complex sentences from simple parts and is important for conveying meaning.  For example, the phrase structure of the English sentence ``The old man the boat.'' is critical for correctly determining its meaning (\autoref{fig:different-parses}).

Transformer models~\citep{Transformer} are state-of-the-art across a wide variety of NLP tasks \citep{bert}, and pretrained Transformers have been shown to learn hierarchical structures after pretraining on large amounts of data \citep{lin2019open,rogers2020primer}.  However, Transformers do not have a hierarchical structure built into the architecture---that is, they don't have an inductive bias toward hierarchical structure \citep{tran2018importance}.  Additionally, Transformers are shown not to perform well on some compositional generalization tasks that require nested structure \citep{cognition}.

We demonstrate that incorporating an inductive bias toward the hierarchical structure of language improves the performance of the Transformer on downstream tasks. %, by adding a layer with an architectural bias toward phrase structure.
We show that this improves compositional generalization and greatly improves translation of predicated argument structure in machine translation.
%There are many possible ways to do this\jmf{check to see if there are any others, probably in the COG space}, and we experiment with adding an architectural bias toward phrase structure\jmf{remove?}. %Specifically we add an extra layer designed for modeling hierarchical phrases to the Transformer.
Specifically, we augment the Transformer to make it more compositional by adding a tree-encoder layer designed for modeling hierarchical phrases.  Additionally we show this layer improves downstream performance across a wide variety of tasks.
%and demonstrate it improves the compositional generalization performance of the Transformer while also improving downstream performance across a variety of tasks.

%\nilay{move this down & reword}
%Prior work on unsupervised syntactic parsing has experimented with neural models for phrase structure \citep{diora,xu2021improved}\jmf{perhaps cite others}.  However, this prior work has focused on unsupervised syntactic parsing and did not explore incorporating a phrase-structure inductive bias into the Transformer for supervised tasks, which is our focus here\jmf{not a general-purpose architecture, specific to unsupervised parsing}.

Our inductive bias layer, which we call \textbf{Treeformer}, is an encoder module that constructs hierarchical phrase encodings and is inspired by the CKY context-free-grammar parsing algorithm \citep{cocke, younger, kasami}. To the best of our knowledge, this is the first study of adding a CKY-style phrase-structure inductive bias into a Transformer for compositional generalization and general-purpose supervised learning.

Prior work has used a similar CKY-style neural architecture for modeling unsupervised syntactic parsing \citep{diora, xu2021improved}. These models are specific to unsupervised parsing and not directly applicable to supervised methods. In contrast, we focus on creating such an architecture for general-purpose supervised learning. Treeformer is also simpler than similar work such as DIORA \citep{diora}, and faster due to two key optimizations which improve the complexity from cubic to linear time (see \autoref{sec:complexity}).

%=====================
%Starting with tokens (i.e., phrases of length 1), we iteratively construct longer phrase encodings by composing subphrases. While encoding a full tree in this way has a run-time complexity of $O(n^{3})$, we can improve this to $O(nH)$ with parallelization and a bound $H$ on the phrase length.

% We demonstrate the effectiveness of adding a Treeformer module to the vanilla Transformer with extensive experiments in machine translation \cite{iwslt}, abstractive summarization \citep{gigaword-dataset, gigaword-rush}, and several natural language understanding tasks \citep{glue}. Additionally, we show the Treeformer module enhances the compositional generalization of a vanilla Transformer on the CoGnition dataset \citep{cognition}, a challenging machine translation dataset for testing compositional generalization.

We demonstrate the effectiveness of adding a Treeformer module to the vanilla Transformer with experiments in compositional generalization (CG) on COGS \cite{COGS} and CoGnition, \cite{cognition}, two challenging seq2seq datasets for testing CG. In addition, the addition of a Treeformer shows significant improvements in machine translation \cite{iwslt}, abstractive summarization \cite{gigaword-dataset, gigaword-rush}, and tasks in natural language understanding \cite{glue-tasks}.  Significantly, we find that the Treeformer is much better at correctly translating predicate-argument structures (subjects vs objects, etc).  Predicate-argument structures require understanding the hierarchical structure of language, and are very important for correctly conveying meaning. This demonstrating the benefits of the Treeformer architecture.

We leave to future work large-scale pretraining with our architecture. While interesting and important for practical considerations, pretraining is not within our computing budget and we consider it out of scope for this work. Our focus is on advancements purely in model architecture.

The paper is organized as follows. First, we discuss some related work (\autoref{sec:related-work}). Then we present our Treeformer module (\autoref{sec:Treeformer}). We analyze the computational complexity and propose two methods for optimizing the algorithm  (\autoref{sec:complexity}). After describing our experimental setups (\autoref{sec:experiments}), we present our results (\autoref{sec:results}) and finally conclude (\autoref{sec:conclusion}).

\section{Related Work}\label{sec:related-work}

% \begin{figure}[ht]
%   \centering
%   \includegraphics[width=0.5\textwidth]{images/tree-lstm-composition.png}
%   \caption{An example of Tree-LSTM composing two children (with subscript 2 and 3) with the parent's
%     state ($c_{1}$ and $h_{1}$). Figure from \citet{treelstm}.}%
%   \label{fig:treelstm}
% \end{figure}
\jmf{check for newer related work}
There is much prior work that induces, operates over, or otherwise uses a tree structure in neural network models \citep{socher-et-al,treelstm,le-and-zuidema,dyer-etal-2016-recurrent, latent-tree-attn, gumbel-tree-lstm,choi2018, diora,only-need-attn, tree-Transformer,recursive-self-attn,r2d2, yogatoma-et-al, Transformer-grammars}. Such models are especially of interest due to the prevalence of trees in natural language.

\citet{treelstm} introduced Tree-LSTMs, an LSTM model generalized to work on parse trees. They suggest specific instances of the general Tree-LSTM architecture for particular types of trees such as dependency and constituency trees. However, Tree-LSTMs and many other tree- or graph-structured models \citep{Nguyen2020Tree-Structured, wang2022learning, tree-positional-encodings, harer2019treeTransformer, Transformer-grammars} require a parse tree over the input text, making data expensive or difficult to obtain. Unsupervised parsing methods \cite{unsupervised-treelstm, tree-Transformer, headsup, diora} have been of interest to solve this problem, but mostly focus on parsing rather than downstream tasks as we do in this paper. One exception is the Gumbel Tree-LSTM \citet{gumbel-tree-lstm}, which use an unsupervised method to generate tree structures for classification tasks. The authors showed improvement on two tasks \cite{snli, sst2} at the time of writing, but they fall short of modern methods such as finetuning pretrained language models.

%\paragraph{Unsupervised Tree-LSTM}
% There is also prior work that induces a tree structure for use in neural models. \citet{unsupervised-treelstm} induces a parse tree using a CKY-style parsing algorithm. Instead of computing the most likely parse, they compute and weigh each possible parse based on cosine similarity to a learned vector. The authors test on two classification tasks: textual entailment \citep{snli} and reverse dictionary \citep{reverse-dictionary}.
% In Tree Transformer \cite{tree-Transformer}, the authors constrain the self-attention module in a Transformer encoder to induce a tree structure by bucketing tokens into groups, disallowing attention across group boundaries, and iteratively merging groups at particular points. This results in tree structure with branches at each merge point which can be used to induce parse trees. A different approach was proposed by \citet{headsup}, in which attention heads in pretrained language models are ranked by intrinsic properties and are shown to be capable of learning a tree structure without supervision.

% Phrase Transformers \cite{phrase-Transformer} improves performance on semantic parsing and machine translation by incorporating $n$-gram models to attention heads in a Transformer, further supporting the case for including hierarchical information.

Most similar to our architecture is the work of \citet{diora}, who introduced Deep Inside-Outside Recursive Autoencoders (DIORA). DIORA learns tree structures using a modified inside-outside algorithm. The inside pass recursively generates a single root node, and the outside pass regenerates the leaf nodes from a root.

DIORA focuses on unsupervised parse tree induction and demonstrates a number of trees that closely match traditionally labeled ones, suggesting the composition algorithm learns efficacious information---a fact we rely on in this paper.
Our Treeformer layer is similar to DIORA's ``inside'' pass but simpler and faster (see \autoref{subsec:algorithm}). Treeformer also has no ``outside'' pass as it does not need to regenerate the leaf nodes, but instead uses the encoded tree structure from the inside pass directly for downstream tasks.

\section{Treeformer}\label{sec:Treeformer}

\begin{figure*}[t]
  \centering
  \includegraphics[width=\textwidth]{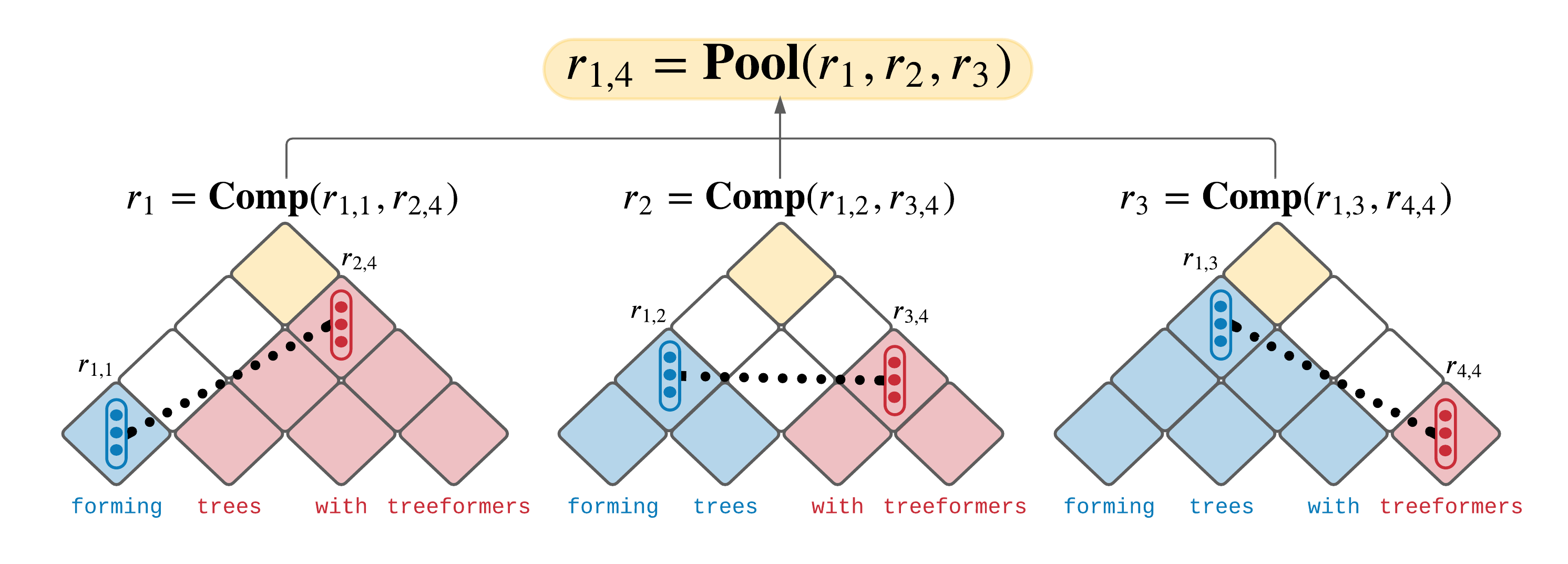}%
  \caption{A demonstration of how the phrase ``forming trees with treeformers'' is encoded. First, we consider each pair of complementary subphrases (each chart represents a different pair). Next, for each pair, we compose their representations using a composition function $\mathbf{Comp}$ into an intermediate representation $r_{k}$. Finally, we pool the intermediate representations into a single vector via some function $\mathbf{Pool}$.}%
  \label{fig:treeformer}
\end{figure*}

The Treeformer algorithm generates phrase encodings by the repeated composition of a given set of token encodings. We start with $n$ tokens (i.e., phrases of length 1) and their representations. We recursively apply the algorithm to compute representations of phrases of length k for all lengths $k$ where $k \leq n$. Our approach, shown in \autoref{fig:treeformer} and \autoref{alg:treeformer}, is inspired by the CKY algorithm. 

\algrenewcommand\algorithmicrequire{\textbf{Input:}}
\algrenewcommand\algorithmicensure{\textbf{Output:}}
\begin{algorithm}
  \caption{Treeformer algorithm}\label{alg:treeformer}
  \begin{algorithmic}[1]
    \Require{$s_{i,j}$, $\{r_{k,k}\ :\ \forall k, i \leq k \leq j\} $} \Comment{Token encodings}
    \Ensure{$r_{i,j}$ }
    \Function{FormTree}{$s_{i,j}$}
    \If{$i = j$} \Comment{Base case}
    \State \textbf{return} $r_{i,j}$
    \EndIf%
    \For{$k \gets i$ to $j$}
    \State $r_{i,k} \gets $ \Call{FormTree}{$s_{i,k}$} \Comment{Recurse}
    \State $r_{k+1,j} \gets $ \Call{FormTree}{$s_{k+1,j}$}
    \State $r_{k} \gets $ \Call{Comp}{$r_{i,k}, r_{k+1,j}$} \Comment{Compose}
    \EndFor
    \State $r_{i,j}\gets$ \Call{Pool}{$r_{i}, \ldots r_{j}$} \Comment{Pool}
    \State \textbf{return} $r_{i,j}$
    \EndFunction
  \end{algorithmic}

\end{algorithm}

\subsection{Notation}\label{subsec:notation}
We now define some notation used throughout the rest of this paper. For input text $s$, let $s_{i,j}$ indicate the span of tokens starting at index $i$ and ending at index $j$ (inclusive), and let $r_{i,j}$ to be the constructed representation of the span $s_{i,j}$. Finally, we use ``phrase'' and ``span'' interchangeably.
%For clarity, we define some notation used throughout the rest of this paper:
%\begin{itemize}
%  \item For input text $s$, let $s_{i,j}$ indicate the span of tokens starting at index $i$ and ending at index $j$ (inclusive). Furthermore, we use the term ``phrase'' and ``span'' interchangeably.
%  \item Let $r_{i,j}$ be the representation constructed by the Treeformer algorithm of the span $s_{i,j}$.
%\end{itemize}

\subsection{Algorithm}\label{subsec:algorithm}%

At a high level, our algorithm works as follows.  The representation of a phrase is constructed by pooling representations of pairs of sub-phrases (see \autoref{fig:treeformer}).  To build the representation of the phrase $s_{i,j}$, we consider all possible pairs of sub-phrases (\textbf{Collect children}), build a representation for each pair using a composition function (\textbf{Compose}), and finally pool these representations into one using an attention-based pooling operation (\textbf{Pool}).

More precisely, given a phrase $s_{i,j}$ of length $n = j - i$, we want to calculate the representation $r_{i,j}$ from its constituent subphrases. \autoref{fig:treeformer} overviews our approach.

\paragraph{Collect children} First, we gather each pair of complementary subphrases of $s_{i,j}$. For each index $k$ such that $i \leq k < j$, we can split $s_{i,j}$ into a pair of subphrases $s_{i,k}$ (prefix) and $s_{k+1,j}$ (suffix). Let $R_{i,j}$ be the set containing the representations of each such pair:
\begin{equation*}
  R_{i,j} = \{(r_{i,k}, r_{k+1,j}) \ :\  i \leq k < j \}
\end{equation*}
\begin{figure}[t]
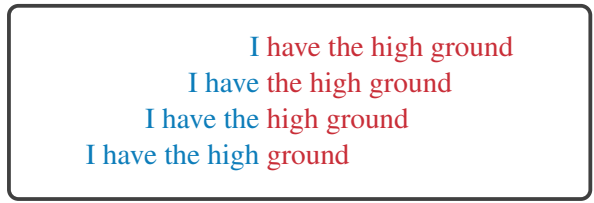

  \begin{tcolorbox}[minipage,arc=0pt, standard jigsaw, opacityback=0, outer arc=0pt, size=normal]
    \centering
    \begin{tabular}{@{}r@{}l@{}}
      \textcolor{mblue}{I }               & \textcolor{mred}{have the high ground} \\
      \textcolor{mblue}{I have }          & \textcolor{mred}{the high ground}      \\
      \textcolor{mblue}{I have the }      & \textcolor{mred}{high ground}          \\
      \textcolor{mblue}{I have the high } & \textcolor{mred}{ground}               \\
    \end{tabular}
  \end{tcolorbox}
  \caption{All \textcolor{mblue}{prefix} and \textcolor{mred}{suffix} pairs of the phrase ``I have
    the high ground''. We might guess the split ``I have'' and ``the high ground'' is the correct parse,
    but the model considers a weighted average of all parses.}\label{fig:example-subphrases}
\end{figure}
\autoref{fig:example-subphrases} shows the four such pairs of the input sentence
${s_{1,5} = \text{``I have the high ground''}}$.  Note that these are exactly the set of pairs we
would consider when parsing with the CKY algorithm.

\paragraph{Compose} Next, we construct a set $C_{i,j}$ as the image of a \textit{composition function}
$\mathbf{Comp} : \mathbb{R}^{d}  \times \mathbb{R}^{d} \rightarrow \mathbb{R}^{d}$ on $R_{i,j}$. That is, it takes \textit{pairs} of vectors and composes them into a
single vector representing the concatenated span:
\begin{equation*}
   C_{i,j} = \{\mathbf{Comp}(r_{k})\ :\ r_{k} \in R_{i,j}\}
\end{equation*}
\normalsize
Because the order of words and phrases in language matters, we want to retain non-commutativity, so this composition function should be non-commutative. A simple example would be concatenating the pair of vectors and feeding the result through a linear transformation. Indeed, Treeformer's composition function is exactly that:
\begin{equation}
  \label{eq:composition-func}
  \mathbf{Comp}(r_{i,k}, r_{k+1,j}) = \mathbf{W} \cdot \left[r_{i,k}, r_{k+1,j}\right]
\end{equation}
where $\mathbf{W} \in \mathbb{R}^{2d\times d}$ and $[\cdot, \cdot]$ indicates concatenation.
Thinking in terms of the CKY algorithm, composing two representations with $\mathbf{Comp}$ is the analogue of applying a grammatical rule.

\paragraph{Pool} Finally, we pool the set $C_{i,j}$ into a single output vector $r_{i,j}$ via some \textit{pooling function} $\mathbf{Pool}$. A simple example would be an average or sum of the vectors, though these options treat all possible parses as equally valid.  Treeformer's pooling function utilizes attention and a model parameter $w \in \mathbb{R}^{d}$.  We calculate a weighted average of each $c_{k} \in C_{i,j}$ using scaled dot-product attention to $w$:

\begin{equation}
  \label{eq:pooling-func}
  r_{i,j} = \sum_{c_{k} \in C_{i,j}} \text{softmax}\left(\frac{\mathbf{K}c_{k} \cdot \mathbf{Q}w}{\sqrt{d}} \right) c_{k}
\end{equation}
At this point in the CKY algorithm, we'd be able to precisely determine our set of valid pairs and eliminate the others using the non-terminals and allowable grammar rules. However, it's not so straightforward to do so with untyped, approximate representations such as vectors. The pooling function is meant do so by extracting only pertinent information from each pair of nodes, each of which represents a possible parse.

\paragraph{Use in Downstream Tasks}
For seq2seq tasks, inserting the Treeformer module is simple. We feed the output of the encoder into the Treeformer, and use the result as the memory for cross attention in the decoder. For sequence classification tasks, we average the top row of the Treeformer output and add the result to the \texttt{[CLS]} token representation from the pretrained Transformer (e.g., ALBERT).
\jmf{add section on how we do seq2seq}

\paragraph{Comparison to DIORA} It is useful to compare the Treeformer architecture to DIORA's inside pass \cite{diora}.  DIORA uses a Tree-LSTM or MLP as the composition function, which we simplify to concatenation followed by a linear projection, which is equivalent to two linear projections added together.  This is faster to compute because the linear projections can be precomputed in $O(n)$ and reused, rather than the $O(n^2)$ computations for DIORA.  Additionally, our pooling function is simplified when compared to DIORA's bilinear compatibility function, which allows us to use linearity to precompute the majority of the computationally expensive operations in our pooling function in $O(n)$ time rather than $O(n^2)$ for DIORA's compatibility function.

\section{Parallelization}\label{sec:complexity}%
\jmf{rename to parallelization or something}

\jmf{In the future, is there a way to demonstrate the empirical complexity of our algorithm? (graph of runtime vs input sentence length)?}
The CKY algorithm, which uses a similar chart structure to Treeformer, has a worst case run-time complexity of $\OO(n^{3} |G|)$ where $|G|$ is the size of the context-free grammar. Similarly, the Treeformer encoding algorithm is also $\OO(n^{3})$ assuming constant model dimension and sequential operations. In this section, we show this calculation as well as two key optimizations which are necessary for tractable training and improve the time and space complexity to $\OO(n)$ and $\OO(nmH)$ respectively. See \autoref{subsec:speed-comparison} for empirical results.

\paragraph{Sequential Algorithm} Starting with a sequence of length $n$, we encode phrases of length $h$ for $1 \leq h \leq n$. There are $n-h+1$ phrases of length $h$, each having $h-1$ pairs of children. Each pair will be composed together exactly once in the entire algorithm, giving us
\begin{equation}\label{eq:base-complexity}
  \sum_{h=1}^{n} (n-h+1)(h-1) = \OO(n^{3})
\end{equation}
total compositions. As our composition function runs in constant time (with respect to $n$), our total complexity for compositions is $\OO(n^{3})$. For pooling, we have $\OO(n^{2})$ total nodes each with $\OO(n)$ pairs of children each. Since the scaled dot-product attention scales linearly in its arguments, we again get a complexity of $\OO(n^{3})$ for pooling, and thus for the entire algorithm as well.

\paragraph{Parallel Algorithm} While encoding phrases of length $h$ is dependent on the encodings for all lengths less than $h$, there is no dependency on other phrases of the same length, allowing us to compute them in parallel. Parallelization removes the factor of $n-h+1$ in \autoref{eq:base-complexity}, leaving
\begin{equation}\label{eq:parallel-complexity}
  \sum_{h=1}^{n} (h-1) = \OO(n^{2})
\end{equation}
total compositions. Likewise, we can pool $\OO(n)$ sets of children in parallel, reducing the pooling (and thus overall) parallel complexity to $\OO(n^{2})$.

\paragraph{Limiting Tree Height} In practice the space complexity turns out to be a bottleneck. Decoding involves calculating and storing cross attention to $\OO(n^{2})$ vectors (compared to $\OO(n)$ for Transformers) for each of the $m$ tokens in the output, resulting in a space complexity of $\OO(n^{2} m)$. To reduce this, we introduce a hyperparameter $H$ which limits the maximum tree height (or phrase length). This results in $\OO(n)$ and $\OO(nmH)$ complexities respectively. Surprisingly, this optimization is not harmful to the model's effectiveness and is possibly even beneficial (see appendix). We find a value of $H=10$ gives the best performance in general, so we use that for all experiments.

\section{Experiments}\label{sec:experiments}%
\jmf{move to appendix}

We conduct experiments in five settings: (1) English-Chinese machine translation for CG on CoGnition \citep{cognition}, (2) semantic parsing for CG on COGS \cite{COGS}, (3) machine translation on IWSLT'14 German-English and English-French \citep{iwslt}, (4) abstractive summarization on GigaWord English abstractive summarization \citep{gigaword-dataset}, and (5) five natural language understanding tasks selected from GLUE \cite{glue-tasks}. For full experimental details, see appendix. Models referred to as ``Treeformer'' are a Transformer with a Treeformer module, as described in the last paragraph in \autoref{subsec:algorithm}.

%\paragraph{Models} Our baseline model is a standard Transformer (Fairseq model configuration \texttt{Transformer\_iwslt\_de\_en}). We trained the models for 100,000 and 60,000 updates for De-En and En-Fr respectively. Furthermore, we evaluate the effect of maximum tree height $H$ by training models with $H \in \{5, 10, 15, 20\}$, and compare performance for models with encoder depth 0 to 6 (decoder is constant at 6 layers). Hyperparameters are detailed in \autoref{sec:appendix-experiments}.

%\subsection{Compositional Generalization}

\jmf{make this more explicit, highlight in the intro, move to own section.}

We test our models on two compositional generalization datasets: CoGnition \citep{cognition}, an English-Chinese machine translation dataset designed to test CG abilities, and COGS \cite{COGS}, a semantic parsing dataset. These datasets are specifically designed to test a model's ability generalize compositionally by testing its ability to generalize to novel combinations of predicates and arguments.

\paragraph{A Note About Baselines}
Although there is much prior work on tree structures in deep learning, we are not aware of any prior work using tree structures that is suitable as a baseline for our tasks beyond the Transformer.  Models such as DIORA \cite{diora} and related models are for unsupervised parsing, and not for classification or seq2seq tasks such as the ones we consider here.  Gumbel Tree-LSTMs \cite{gumbel-tree-lstm}, similarly are only for classification, and not for seq2seq. Transformer Grammars \cite{Transformer-grammars} and RNNGs are for parsing or language modeling \cite{dyer-etal-2016-recurrent}, or for classification \cite{yogatoma-et-al}.  All the above architectures would require significant changes for seq2seq tasks.

\begin{table*}[t]
  \centering
  \begin{tabular}{lrrr}
    \toprule
    Model                            & Parameters & De-En       & En-Fr       \\
    \midrule
    Transformer \citep{dynamic-conv} &    37M     & 34.4        & -           \\
    DynamicConv \citep{dynamic-conv} &    -       & 35.2        & -           \\
    \midrule
    \rowcolor{verylightgray}
    Transformer                      & 37M        & 34.5        & 41.0        \\
    \rowcolor{verylightgray}
    Transformer (7-layer encoder)    & 42M        & 34.9        & 41.2        \\
    \rowcolor{verylightgray}
    Treeformer $(H=10)$              & 40M        & \bold{35.4} & \bold{41.5} \\
    \midrule
    BiBERT (state of the art)        &            & 38.6        & -           \\
    \bottomrule
  \end{tabular}
  \caption{Model performance (BLEU) on the IWSLT'14 German-English and English-French translation tasks. Models we trained (highlighted in grey) used six layer encoders and decoders and dimensions $d_{model} = 512 \text{ and } {d_{ffn} = 1024}$. For comparison, we also report the (to the best of our knowledge) state-of-the-art for De-En \citep{bibert}\jmf{maybe add LSTM and gumble-tree softmax results}.}%
  \label{tab:iwslt-results}
\end{table*}

\section{Results}\label{sec:results}%
%We now present and analyze our results.

\subsection{Translation}\label{subsec:results-translation}
\autoref{tab:iwslt-results} shows the results on IWSLT'14 German-English and English-French translation. Compared to the baseline Transformer, our model improves by 0.9 and 0.5 BLEU points over a 6-layer Transformer, and by 0.5 and 0.3 over a Transformer with a 7-layer encoder (which notably has more parameters than the Treeformer). For German-English, we also report scores from DynamicConv \citep{dynamic-conv} and their reported baseline (also a Transformer), compared to which our model improves by 0.2 and 1.0 points respectively.

\begin{figure*}[ht]
\centering
  \includegraphics[width=0.8\textwidth]{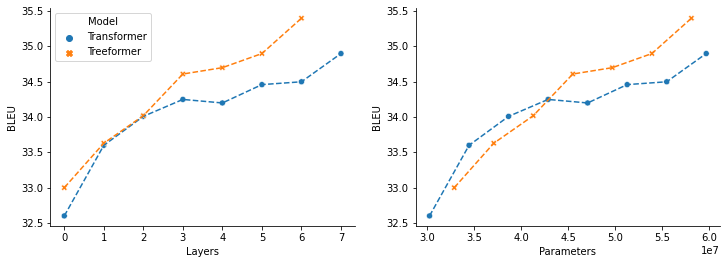}
  \caption{Effects of including a Treeformer module on-top of a Transformer with respect to the number of layers (left) and parameters (right). Although the Treeformer module is less efficient in shallower models, its efficacy grows as the underlying encoder grows larger. With more layers, it becomes more parameter-efficient to add a Treeformer module than adding more Transformer layers. Models are trained and evaluated on IWSLT'14 De-En.}
  \label{fig:effect-of-size}
\end{figure*}

\subsection{Abstractive Summarization}%
\label{subsec:results-abstractive-summarization}
For the summarization task , Treeformer improves by a significant 1.6, 0.9, and 0.6 points in ROUGE-1, ROUGE-2, and ROUGE-L respectively compared to the baseline (\autoref{tab:gigaword-results}). 
\begin{table*}[t]
  \centering
  \begin{tabular}{l r r r r}
    \toprule
    Model                       & Parameters & ROUGE-1     & ROUGE-2     & ROUGE-L     \\
    \midrule
    Transformer                 & 73M & 37.1        & 17.7        & 34.8        \\
    Treeformer ($H = 10$)       & 75M & \bold{38.7} & \bold{18.6} & \bold{35.4} \\
    \midrule
    Pegasus+DotProd (state of the art) &  568M      & 40.45       & 20.69       & 36.56       \\
    \bottomrule
  \end{tabular}
  \caption{Model performance (ROUGE) on the Gigaword abstractive summarization task. Bold values indicate the highest performance for each metric. We include the current (to the best of our knowledge) state-of-the-art \citep{pegasus-dotprod}.\jmf{check still SOTA}}%
  \label{tab:gigaword-results}
\end{table*}

\subsection{GLUE}%
\label{subsec:results-glue}
\begin{table*}[t]
  \centering
  \begin{tabular}{lrrrrrr}
    \toprule
    GLUE Task   & CoLA        & MNLI (m/mm)        & MRPC               & SST-2       & STS-B & Avg.              \\
    \midrule
    ALBERT      & 56.4        & 84.9 / 85.1        & \bold{88.9 / 92.0} & 91.9        & \bold{90.4 / 90.7} & 85.0 \\
    ALBERT+Treeformer & \bold{61.5} & \bold{85.4 / 85.5} & 88.4 / 91.6        & \bold{92.4} & \bold{90.4 / 90.7} & \bold{85.7}\\
    \bottomrule
  \end{tabular}
  \caption{ Model performance on selected GLUE tasks. ALBERT is the \texttt{albert-base-v2} pretrained model from Huggingface's \texttt{Transformers} library, fine-tuned on these five tasks. We add a Treeformer as described in \autoref{subsec:algorithm}}%
  \label{tab:glue-validation-scores}
\end{table*} 

Treeformer matches or improves performance on four of five selected GLUE tasks, notably making a significant improvement on CoLA with a 5.1 point increase (\autoref{tab:glue-validation-scores}). Intuitively, we expect Treeformer to perform well on single-sentence tasks more so than sentence pair tasks since phrases that span both sentences would likely be meaningless. This is reflected in our results as Treeformer performs well on both CoLA and SST-2. These results indicate despite rich contextual token encodings, Transformers are not capturing beneficial phrase-level information.

\subsection{Compositional Generalization}

On the CoGnition CG test set (\autoref{tab:cg-results}), Treeformer attains a significant 4.2\% and 5.9\% decrease in instance-level and aggregate-level compound error rates respectively (averaged over three runs).

On COGS, Treeformer improves over the Transformer by 1.6\% percentage points. Our results on both datasets indicate the hierarchical structure is especially useful for generalization tasks while simultaneously improving other downstream tasks.
\begin{table*}[ht]
  \centering
  \begin{tabular}{lcc}
    \toprule
    Model & CoGnition (Inst/Agg. ER) $\downarrow$ & COGS (Acc.) $\uparrow$  \\
    \midrule
    Transformer & 29.0/64.3\% & 78.5\% \\
    Treeformer & \bold{24.8/58.5\%} & \bold{80.1\%} \\
    \midrule
    T5+CSL-Aug \cite{latent-structure} & - & 99.5\% \\
    R-Dangle \cite{rdangle} & 16.0/42.1\% & - \\
    \bottomrule
  \end{tabular}
  \caption{Results on the CoGnition COGS datasets. In both cases, the Treeformer makes significant improvements in generalization ability. For comparison, we also report state-of-the-art for both tasks.}
  \label{tab:cg-results}
\end{table*}

% \begin{table}[!h]
%   \centering
%   \begin{tabular}{lcc}
%     \toprule
%     & Transformer & Treeformer \\
%     \midrule
%     BLEU & 59.5 & 59.6 \\
%     Inst. ER & 29.0\% & \bold{24.8\%} \\
%     Agg. ER & 64.3\% & \bold{58.4\%} \\
%     \bottomrule
%   \end{tabular}
%   \caption{BLEU score and instance-level and aggregate-level error rates on the CoGnition dataset \citep{cognition} (CG test set) for a 6-layer Transformer and Treeformer. The best result in each category is in bold.}
%   \label{tab:cg-results}
% \end{table}

\begin{figure}[h]
 \centering
 \includegraphics[width=0.4\textwidth]{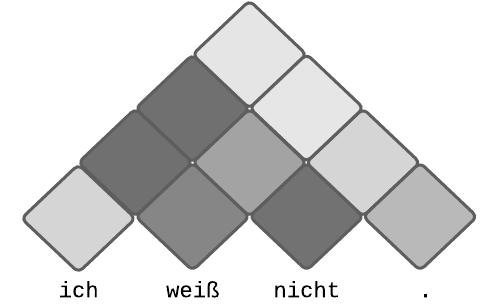}
 \caption{Heat-map of the cross-attention weights for the Treeformer averaged over each layer, head, and output position (darker is higher). \jmf{change to grayscale where darker is higher}\jmf{talk more about this, and say it doesn't always find linguistic trees}}
 \label{fig:heatmap-ich-weiss-nicht}
 \end{figure}

\subsection{Effects of Model Size}
\label{sec:effects-of-model-size}
\autoref{fig:effect-of-size} shows a comparison of the Transformer with and without a Treeformer module at various encoder depths (left) and their respective parameter counts (right). In each case, simply adding a Treeformer module is beneficial, especially in deeper models. Importantly, the Treeformer module becomes more parameter-efficient than further encoder layers as the base model deepens. This fact implies it is not simply extra parameters improving performance, but rather that the Treeformer module is capturing useful information otherwise lost. 

%However, interpreting the nature of this useful information is difficult. 

\subsection{Analysis of Treeformer Attention}
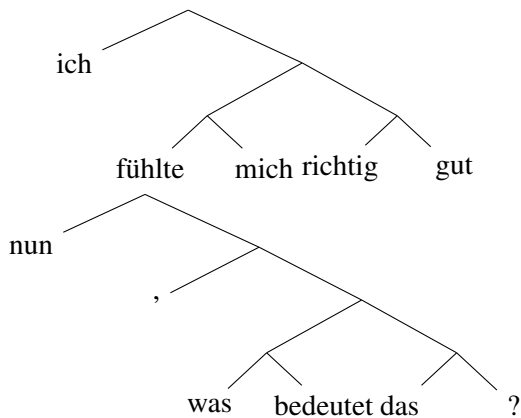
\begin{figure}[t]
  \centering
  \begin{tikzpicture}
    \tikzstyle{level 1}=[sibling distance=30mm, level distance=7mm]
    \tikzstyle{level 2}=[sibling distance=25mm, level distance=7mm]
    \tikzstyle{level 3}=[sibling distance=15mm, level distance=7mm]
    \coordinate (a)
        child {node (b) {ich}}
        child { coordinate (i) 
            child { coordinate (g)
                child {node (c) {f\"{u}hlte}}
                child {node (d) {mich}}
            }
            child { coordinate (h)
                child {node (e) {richtig}}
                child {node (f) {gut}}
            }
        };
  \end{tikzpicture}
  \begin{tikzpicture}
    \tikzstyle{level 1}=[sibling distance=30mm, level distance=7mm]
    \tikzstyle{level 2}=[sibling distance=27mm, level distance=7mm]
    \tikzstyle{level 3}=[sibling distance=25mm, level distance=7mm]
    \tikzstyle{level 4}=[sibling distance=15mm, level distance=7mm]
    \coordinate (a)
        child {node (b) {nun}}
        child { coordinate (j)
            child {node (k) {,}}
            child { coordinate (h) 
                child { coordinate (g)
                    child {node (c) {was}}
                    child {node (d) {bedeutet}}
                }
                child { coordinate (i)
                    child {node (e) {das}}
                    child {node (f) {?}}
                }
            }
        };
  \end{tikzpicture}
  \caption{Example German parses from the model trained on IWSLT'14. Despite no explicit training, the resulting trees are visually plausible.}%
  \label{fig:german-parses}
\end{figure}

In some cases, despite no supervision for parsing, we see the decoder cross-attends to constituent phrases identified by linguists (\autoref{fig:heatmap-ich-weiss-nicht}). Similarly, we can generate ``parse trees'' by choosing the pair with the highest attention weight at each step in the algorithm. In \autoref{fig:german-parses}, we see two such parses which seem visually plausible despite no explicit supervision. However, we find in most cases the generated trees are not linguistically plausible, and do not have high parsing accuracy when evaluated as parse trees. Nevertheless, the improvement in performance we see across tasks, especially for CG and predicate-argument structure in MT, suggests that the information in the phrase-level vectors is useful for understanding the hierarchical structure of language. \nilay{rephrase this section?}

% \begin{table*}[htb]
%     \centering
%     \begin{tabular}{lcccccc}
%         \toprule
%         Model       & Correct & Lexical & Pred-Arg & Morph. & Drop/Add  & Other \\
%         \midrule
%         Transformer & 22 & 25 & 7 & 9 & \bold{3} & 1 \\
%         Treeformer  & \bold{23} & \bold{22} & \bold{2} & \bold{7} & 4 & 1 \\
%         \bottomrule
%     \end{tabular}
%     \caption{Results of a human analysis of 50 randomly sampled sentences from IWSLT'14 De/En, categorized by translation error type. In particular, Treeformer greatly reduces errors in predicate-argument structures.}
%     \label{tab:human-analysis}
% \end{table*}

\begin{table}[h]
    \centering
    \begin{tabular}{lcc}
        \toprule
        Model    & Transformer & Treeformer \\
        \midrule
        Correct  & 22          & \bold{23}         \\
        Lexical  & 25          & \bold{22}         \\
        Pred-Arg & 7           & \bold{2}         \\
        Morphosyntax   & 9           & \bold{7}         \\
        Drop/Add & \bold{3}    & 4          \\
        Other    & 1           & 1          \\
        \midrule 
        Total Errors    & 42          & \bold{34} \\
        \bottomrule
    \end{tabular}
    \caption{Counts from a human analysis of 50 randomly sampled sentences from IWSLT'14 De/En, categorized by translation error type.  The Treeformer greatly reduces errors in predicate-argument structures, demonstrating the benefit of modeling hierarchical structure. The error types are: correct = correct translation, lexical = incorrect lexical choice, pred-arg = incorrect predicate-argument structure (e.g., swapping subjects and objects), morphosyntax = morphosyntactic errors (e.g., incorrect inflections, tense, number, or determiners), drop/add = missing or incorrectly added tokens, other = other errors. Note: sentences can have multiple errors.}
    \label{tab:human-analysis}
\end{table}

%This hypothesis is further supported by the attention heatmap in \autoref{fig:heatmap-ich-weiss-nicht} which shows that the decoder attends to constituent phrases identified by linguists such as "ich weiß" and "ich weiß nicht", suggesting that the Treeformer is learning a hierarchical linguistic structure.

%However, this is not true for shallow models, indicating that sufficiently rich token encodings are necessary before phrase encodings make a meaningful difference. 

\subsection{Treeformer Captures Predicate-Argument Structure}
\label{subsec:human-analysis}

To better understand where Treeformer improves over a vanilla Transformer, we conduct a human analysis on 50 randomly selected examples from the IWSLT'14 De/En validation set (\autoref{tab:human-analysis}). We find the Treeformer greatly reduces the frequency of errors in predicate-argument structure (e.g., swapping subject and object, or the example in \autoref{tab:iwslt-example}).  Of the categories of errors we analyzed, correctly translating predicate-argument structure requires the most understanding of the hierarchical structure, and is very important for correctly conveying the meaning. This demonstrates the benefit of the Treeformer approach.

\begin{table}[htb]
    \centering
    \begin{tabular}{ll}
        \toprule
        Input & also ging ich von da an weiter. \\
        Transformer & so i went from there to further. \\
        Treeformer & so i went on from there. \\
        Gold & so i moved on from there. \\
        \bottomrule
    \end{tabular}
    \caption{An example from the IWSLT'14 validation set in which the vanilla Transformer makes a predicate-argument error which the addition of the Treeformer avoids.}
    \label{tab:iwslt-example}
\end{table}

\subsection{Speed Comparison}
\label{subsec:speed-comparison}

\begin{figure}[!ht]
    \centering
    \includegraphics[width=0.4\textwidth]{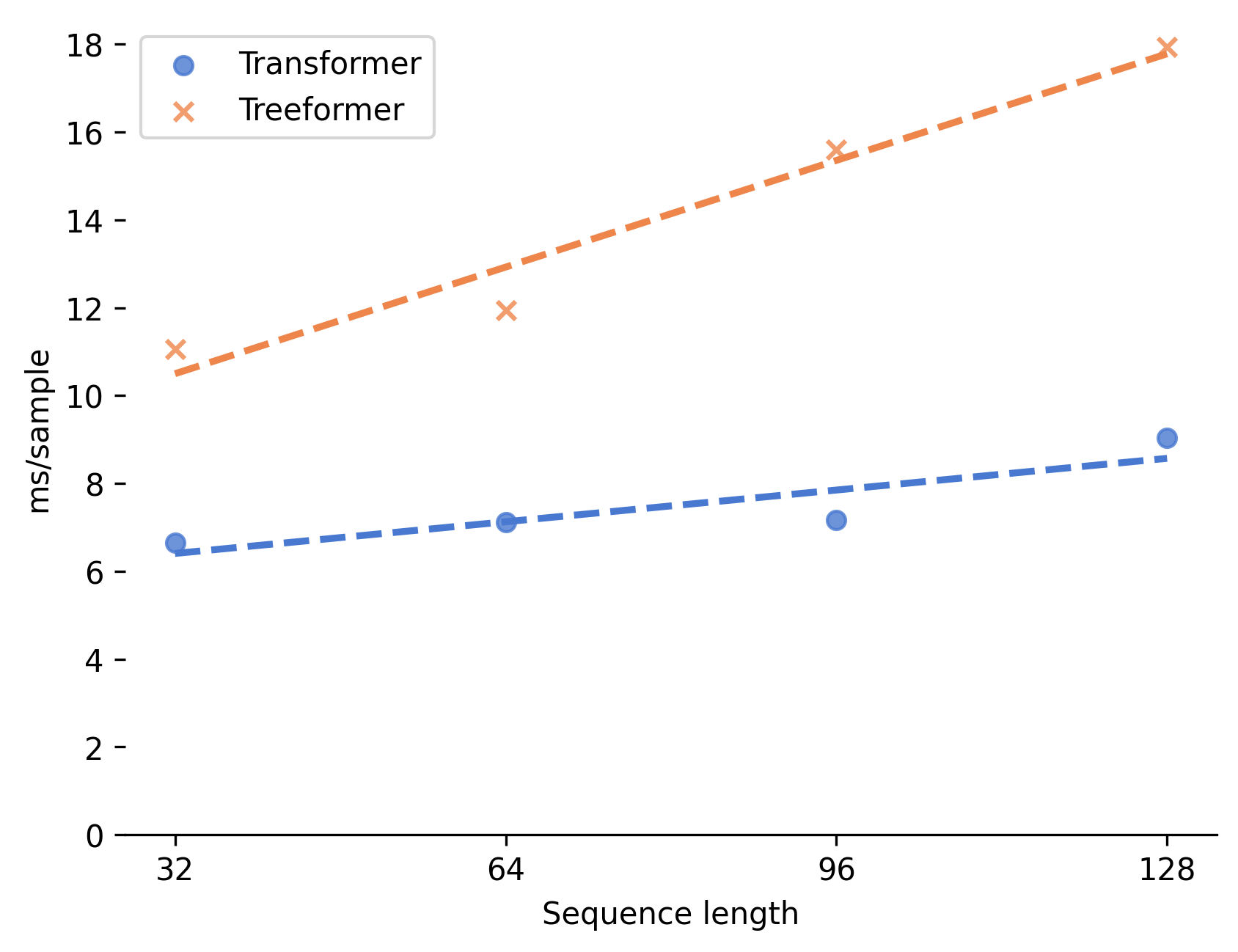}
    \caption{A comparison of training speed (ms/sample) by sequence length. The Treeformer is about 50\%-60\% as fast as the Transformer. For shorter sequences, Treeformer is about 60\% as fast, which decreases to about 50\% for longer sequences.}
    \label{fig:speed-comparison}
\end{figure}

Our optimizations (\autoref{sec:complexity}) make training Treeformer tractable, but the architecture is slower than the vanilla Transformer due to the sequential nature of the algorithm and the increase in total encoded vectors. We measure the encoder-only speed at various sequence lengths for both models (\autoref{fig:speed-comparison}).

\section{Conclusion}%
\label{sec:conclusion}

This paper presents Treeformer, a CKY-inspired neural network algorithm for composing tokens into phrases and sentences. We showed that, in many cases, standard Transformers are unable to effectively capture the phrase-level or hierarchical information which the Treeformer module helps exploit. This information allows the Treeformer to outperform a vanilla Transformer in compositional generalization and many downstream tasks including machine translation, abstractive summarization, and natural language understanding.

We believe hierarchical structure is an important feature for models to have due to the prevalence of tree structures in natural language, and are further convinced by the performance increase shown with our Treeformer module across a variety of settings. While this paper and many previous works modify algorithms such as CKY to induce tree structures, this approach can be slow and resource intensive due to the number of parses which must be computed. We believe improving speed, memory, and performance in tree-level neural models is possible and an important avenue for future research.

%this section is mandatated by EACL, but doesn't count towards the page limit
%\section*{Limitations}
%One possible limitation of our proposed architecture is its speed and memory requirements. With our optimizations and with $H=10$, our model is approximately twice as slow as a standard Transformer. This limitation may become relevant when considering massive datasets such as those used in pretraining large language models.

\bibliography{treeformers}
\bibliographystyle{acl_natbib}

\clearpage
%\begin{comment}
    
\appendix
\section{Experimental Setup}\label{sec:appendix-experiments}%
All code is written in PyTorch \citep{pytorch}. For both seq2seq tasks, we use Fairseq \citep{fairseq},\footnote{\url{https://github.com/pytorch/fairseq}} a sequence modeling framework from Facebook AI, for preprocessing and training. All models are trained on a single NVIDIA GeForce RTX 3090. Hyperparameters are detailed in \autoref{tab:iwslt-hyperparams} and \autoref{tab:albert-glue-hparams}.

\begin{table*}[htb]
    \centering
    \begin{tabular}{lccc}
    \toprule
    Dataset & Train/Val/Test Size & Approx. Training Time & Evaluation \\
    \midrule
    IWSLT'14 De/En & 168K/-/7K & 12h & BLEU \\
    IWSLT'14 Fr/En & 186K/-/9K & 12h & BLEU \\
    Gigaword & 4M/189K/2K & 24h & ROUGE-\{1,2,L\} \\
    CoGnition & 196K/10K/10,11K & 5h & BLEU/Error Rate \\
    COGS & 24K/3K/21K & 2h & Accuracy \\
    \bottomrule
    \end{tabular}
    \caption{Dataset statistics}
    \label{tab:dataset-stats}
\end{table*}

\paragraph{Compositional Generalization}
For CoGnition, we use the provided training script without hyperparameter modification\footnote{\url{https://github.com/yafuly/CoGnition}}. For COGS, we use the training scheme suggested by \citet{devil} re-implemented in Fairseq. Our results do not exactly reproduce theirs, but are consistently 1.5\% off in each of their settings for COGS. Therefore we believe the improvement from Treeformer is still sound and not due to any improper experimental design.

\paragraph{Machine Translation}\label{subsec:machine-translation}
Our machine translation experiments are on IWSLT'14 German-English and English-French translation tasks \citep{iwslt}. We use the recommended hyperparameters, as nothing better was found in our limited search. We use the Moses tokenizer\footnote{\url{https://github.com/moses-smt/mosesdecoder}}\citep{moses-tokenizer} and Subword NMT\footnote{\url{https://github.com/rsennrich/subword-nmt}}\citep{subword-nmt} for byte-pair encoding. We report BLEU scores \citep{bleu} (calculated using Fairseq) on the test set for the best models of each configuration, selected by validation performance.

\paragraph{Abstractive Summarization}\label{subsec:abstractive-summarization}
We test our models on the Gigaword abstractive summarization dataset \citep{gigaword-dataset, gigaword-rush}, a corpus of nearly 4 million document-headline pairs in which the model is meant to summarize the document into its corresponding headline \autoref{tab:dataset-stats}. We use the same two model architectures as in the machine translation experiments described above. For the Treeformer, we use a tree height of $H=10$. Hyperparameters are detailed in \autoref{tab:iwslt-hyperparams}. The models are evaluated on the Gigaword test set on ROUGE-1, -2, and -L scores\citep{rouge}, calculated with the python package \texttt{rouge}\footnote{\url{https://github.com/pltrdy/rouge}}.

\paragraph{GLUE} 
For GLUE, we fine-tune a pretrained ALBERT base model \citep{albert} from on five of the GLUE tasks\footnote{These were chosen to minimize training time while also providing variety.} \citep{cola, sst2, stsb, mnli, mrpc}. Due to accessibility, we report scores on the provided validation set and instead hold out 10\% of the training data for model selection.  We select five GLUE tasks to conduct our experiments on sequence classification: Corpus of Linguistic Acceptability (CoLA) \citep{cola}, Multi-NLI \citep{mnli}, Microsoft Research Paraphrase Corpus (MRPC) \citep{mrpc}, Stanford Sentiment Treebank (SST-2) \citep{sst2}, and Semantic Textual Similarity Benchmark (STS-B) \citep{stsb}.For each task, we use ALBERT (\texttt{albert-base-v2}) from huggingface's \texttt{Transformers} both with and without a Treeformer module. The evaluation metrics are the standard ones for GLUE.

\section{Effect of Maximum Height}%
\label{sec:results-effect-of-max-height}

\begin{figure}[H]
  \centering
  \includegraphics[width=0.45\textwidth]{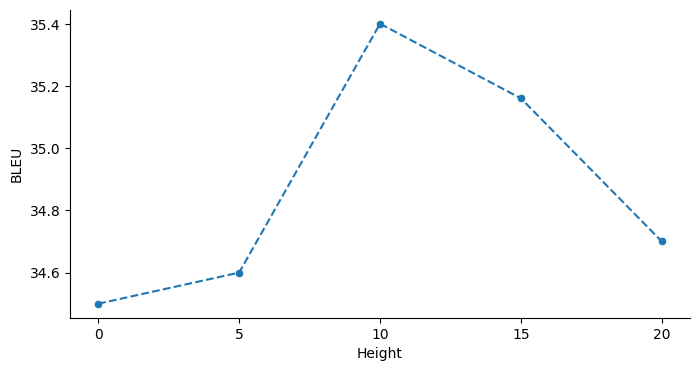}
  \caption{The effect of tree height on Treeformer BLEU score on IWSLT'14 ($H=0$ is a vanilla Transformer). There appears to be a sweet spot around $H \approx 10$ for which longer phrases are no longer beneficial.}
  \label{fig:comparison-of-height}
\end{figure}

We investigate the effect of the maximum tree height $H$ in the Treeformer (\autoref{fig:comparison-of-height}). Interestingly, we find that increasing the maximum tree height only improves performance up to a certain point, which we find to be near 10. We hypothesize that long phrases (over 10 tokens) typically aren't as salient when the full sentences are only 30 to 40 tokens. Experiments on GLUE tasks also indicated that $H=10$ is near optimal, supporting this hypothesis.

Another possibility is the inability of fixed-dimension vectors to capture the information in a composition while retaining information about its important constituents. Sentences, in particular, contain strictly more information than a single token but are equivalently represented in our model (and many others). We leave this investigation for future research.

%\begin{table}[H]
%  \centering
%  \begin{tabular}{ll}
%    \toprule
%    Task  & Metric             \\
%    \midrule
%    CoLA  & Matthew's Corr     \\
%    MNLI  & Accuracy           \\
%    MRPC  & F1/Accuracy        \\
%    SST-2 & Accuracy           \\
%    STS-B & Pearson/Spearman R \\
%    \bottomrule
%  \end{tabular}
%  \caption{Metrics for each GLUE task used in this paper.}%
%  \label{tab:glue-metrics}
%\end{table}

\begin{table*}[h]
  \centering
  \begin{tabular}{l c c c c c}
    \toprule
                    & \multicolumn{2}{c}{IWSLT'14 (De-En, En-Fr)} 
                    & & \multicolumn{2}{c}{Gigaword} \\
    \cmidrule{2-3}
    \cmidrule{5-6}
    Parameter       & Transformer                                 & Treeformer                   & & Transformer              & Treeformer               \\
    \midrule 
    vocabulary     & \multicolumn{2}{c}{8844/6628 (de/en)}                                       & & \multicolumn{2}{c}{32,000}\\
    $d_{model}$     & 512                                         & 512                          & & 512                      & 512                      \\
    $d_{ffn}$       & 1024                                        & 1024                         & & 1024                     & 1024                     \\
    Tokens/batch    & 4096                                        & 4096                         & & 8192                     & 8192                     \\
    Learning rate   & 5e-04                                       & 7e-04                        & & 5e-04                    & 7e-04                    \\
    LR Scheduler    & $\textit{inverse\_sqrt}$                    & $\textit{inverse\_sqrt}$     & & $\textit{inverse\_sqrt}$ & $\textit{inverse\_sqrt}$ \\
    Warmup steps    & 4000                                        & 4000                         & & 8000                     & 8000                     \\
    Dropout         & 0.3                                         & 0.4                          & & 0.1                      & 0.1                      \\
    Weight decay    & 1e-4                                        & 1e-4                         & & 1e-3                     & 1e-3                     \\
    Label smoothing & 0.1                                         & 0.1                          & & 0.1                      & 0.1                      \\
    Beam width      & 10                                          & 10                           & & 5                        & 5                        \\
    Length penalty  & 0.6 (De)/0.2 (Fr)                           & 0.6 (De)/0.2 (Fr)            & & 0.7                      & 0.7                      \\
    Max output length  & -                                        & -                            & & 32                       & 32                      \\
    \bottomrule
  \end{tabular}
  \caption{Fairseq hyperparameters used for training on the IWSLT'14 De-En/En-Fr and Gigaword
    Abstractive Summarization tasks.}%
  \label{tab:iwslt-hyperparams}
\end{table*}

\begin{table*}[h]
  \centering
-
  \begin{tabular}{c l r r r r r }
    \toprule
                        & Parameter          & CoLA              & MNLI              & MRPC              & SST-2             & STS-B             \\
    \midrule
                        & $d_{model}$        & 768               & 768               & 768               & 768               & 768               \\
                        & $d_{ffn}$          & 3072              & 3072              & 3072              & 3072              & 3072              \\
                        & $n_{heads}$        & 12                & 12                & 12                & 12                & 12                \\
                        & Batch size         & 16                & 128               & 32                & 32                & 16                \\
    ALBERT              & Learning rate      & 1e-05             & 3e-05             & 2e-05             & 1e-05             & 2e-05             \\
                        & LR Scheduler       & $\textit{cosine}$ & $\textit{cosine}$ & $\textit{cosine}$ & $\textit{cosine}$ & $\textit{cosine}$ \\
                        & Warm-up steps      & 320               & 1000              & 200               & 1256              & 214               \\
                        & Dropout            & 0.0               & 0.1               & 0.0               & 0.0               & 0.0               \\
                        & Classifier dropout & 0.1               & 0.0               & 0.1               & 0.1               & 0.1               \\
    \midrule
                        & $d_{model}$        & 768               & 768               & 768               & 768               & 768               \\
                        & $d_{ffn}$          & 3072              & 3072              & 3072              & 3072              & 3072              \\
                        & $n_{heads}$        & 12                & 12                & 12                & 12                & 12                \\
                        & Batch size         & 16                & 128               & 16                & 16                & 16                \\
    ALBERT + Treeformer & Learning rate      & 2e-05             & 1.76e-05          & 1.63e-05          & 1e-05             & 3e-05             \\
                        & LR Scheduler       & $\textit{cosine}$ & $\textit{cosine}$ & $\textit{cosine}$ & $\textit{cosine}$ & $\textit{cosine}$ \\
                        & Warmup steps       & 320               & 1000              & 100               & 400               & 642               \\
                        & Dropout            & 0.0               & 0.1               & 0.04              & 0.15              & 0.1               \\
                        & Classifier dropout & 0.1               & 0.0               & 0.08              & 0.1               & 0.1               \\
    \bottomrule
  \end{tabular}
  \caption{Hyperparameters for training ALBERT and ALBERT+Treeformer on each of the GLUE tasks. A random search of 15 trials produced no results better than the recommended ones for ALBERT fine
    tuning \citep{albert}. }%
  \label{tab:albert-glue-hparams}
\end{table*}
\pagebreak

% As shown in \autoref{fig:comparison-of-height}, Treeformer tends to perform optimally around $H=10$, which suggests the presence of very long phrases may contribute too much noise to the decoder. 

%\end{comment}
\end{document}